%% file: sn-article.tex
\documentclass[sn-mathphys,Numbered]{sn-jnl}

\usepackage{graphicx}%
\usepackage{multirow}%
\usepackage{amsmath,amssymb,amsfonts}%
\usepackage{amsthm}%
\usepackage{mathrsfs}%
\usepackage[title]{appendix}%
\usepackage{xcolor}%
\usepackage{textcomp}%
\usepackage{manyfoot}%
\usepackage{booktabs}%
\usepackage{algorithm}%
\usepackage{algorithmicx}%
\usepackage{algpseudocode}%
\usepackage{listings}%

\usepackage{amsthm}
\usepackage{xcolor}
\usepackage[figuresright]{rotating}

\usepackage{graphicx}
\graphicspath{{/}{fig/}}

\usepackage{array}
\usepackage{textcomp}
\usepackage{xcolor}
\usepackage{multirow}
\usepackage{booktabs}

\usepackage{mathtools}
\usepackage{breqn}
\usepackage{float}
\usepackage{subfloat}
\usepackage{subcaption}

\usepackage{pgfplots}
\pgfplotsset{compat=1.7}
\usepgfplotslibrary{groupplots}

\usepackage[font=footnotesize]{caption}
\usepackage{subcaption}

\usepackage{multirow,tabularx}
\usepackage{hyperref}
\usepackage{flushend}
\usepackage[numbers]{natbib}
\usepackage{soul}

\newlength\figureheight
\newlength\figurewidth
\usepackage{adjustbox}

\theoremstyle{thmstyleone}%
\theoremstyle{thmstyletwo}%

\theoremstyle{thmstylethree}%

\begin{document}

\title[DDS, MQTT, and Zenoh for ROS\,2]{Comparison of Middlewares in Edge-to-Edge and Edge-to-Cloud Communication for Distributed ROS\,2 Systems}


\author*[1]{\fnm{Jiaqiang} \sur{Zhang}}\email{jiaqiang.zhang@utu.fi}

\author[1]{\fnm{Xianjia} \sur{Yu}}\email{xianjia.yu@utu.fi}

\author[1]{\fnm{Sier} \sur{Ha}}\email{sierha@utu.fi}

\author[1]{\fnm{Jorge} \sur{Pe\~na Queralta}}\email{jopequ@utu.fi}

\author[1]{\fnm{Tomi} \sur{Westerlund}}\email{tovewe@utu.fi}

\affil*[1]{\orgdiv{Turku Intelligent Embedded and Robotic Systems (TIERS) Lab}, \orgname{University of Turku}, \orgaddress{\city{Turku}, \country{Finland}}}


\abstract{
The increased data transmission and number of devices involved in communications among distributed systems make it challenging yet significantly necessary to have an efficient and reliable networking middleware. In robotics and autonomous systems, the wide application of ROS\,2 brings the possibility of utilizing various networking middlewares together with DDS in ROS\,2 for better communication among edge devices or between edge devices and the cloud. However, there is a lack of comprehensive communication performance comparison of integrating these networking middlewares with ROS\,2. In this study, we provide a quantitative analysis for the communication performance of utilized networking middlewares including MQTT and Zenoh alongside DDS in ROS\,2 among a multiple host system. For a complete and reliable comparison, we calculate the latency and throughput of these middlewares by sending distinct amounts and types of data through different network setups including Ethernet, Wi-Fi, and 4G. To further extend the evaluation to real-world application scenarios, we assess the drift error (the position changes) over time caused by these networking middlewares with the robot moving in an identical square-shaped path. Our results show that CycloneDDS performs better under Ethernet while Zenoh performs better under Wi-Fi and 4G. In the actual robot test, the robot moving trajectory drift error over time (96\,s) via Zenoh is the smallest. It is worth noting we have a discussion of the CPU utilization of these networking middlewares and the performance impact caused by enabling the security feature in ROS\,2 at the end of the paper.
}



\maketitle

\input{sec/01_Intro.tex}

\input{sec/02_Background_and_Concepts.tex}

\input{sec/03_Solutions.tex}

\input{sec/04_Applications.tex}
\input{sec/05_Discussion.tex}

\input{sec/06_Conclusion.tex}

\backmatter




\bmhead{Acknowledgments}

This research work is supported by the R3Swarms project funded by the Secure Systems Research Center (SSRC), Technology Innovation Institute (TII).



\bibliography{bibliography.bib}

\section*{Statements and Declarations}


\begin{itemize}
\item This work was supported by the R3Swarms project funded by the Secure Systems Research Center (SSRC), Technology Innovation Institute (TII). 
\item The authors have no relevant financial or non-financial interests to disclose.
\item No Ethics approval is needed.
\item No Consent to participate is needed.
\item No Consent for publication is needed.
\item All authors contributed to the study conception and design. Material preparation, data collection and analysis were performed by Jiaqiang Zhang, Xianjia Yu and Sier Ha. The first draft of the manuscript was written by Jiaqiang Zhang and all authors commented on previous versions of the manuscript. All authors read and approved the final manuscript.
\end{itemize}






\end{document}

%% file: sec/01_Intro.tex


\section{Introduction}\label{sec:introduction}
Efficient and reliable communication protocols for distributed systems are in significant demand with the development of the Internet of Things (IoT) and edge computing. In addition to the enormous amount of data generated in the system, the heterogeneous devices and communication protocols involved make data sharing and communication among edge devices and the cloud a challenging task. In the field of robotics and autonomous systems, 
the Robot Operating System (ROS) has become the \textit{de-facto} standard. 
The updated version for distributed systems, ROS\,2, has changed the middleware, enabling robots to more effectively utilize the computational resources of the cloud-edge continuum{~\cite{zhang2022distributed}}.

In order to adapt to the trend of decentralized and distributed systems in the field of robotics, ROS\,2 uses Data Distribution Service (DDS) as its internal communication middleware, namely ROS Middleware (RMW) which provides an abstraction layer to different DDS implementations for communication with the ROS 2 Client Library. 
DDS~\cite{pardo2003omg} is a data-centric middleware solution that emphasizes efficient and reliable data transfer between devices. DDS provides a rich set of Quality of Service (QoS) policies that allow applications to specify the reliability, latency, and throughput requirements of their data.
The number of studies and analyses comparing the performance of RMWs is limited~\cite{RMWreports, FastDDSPerformance}.
However, these studies are primarily conducted on a single machine and lack a comprehensive system-level analysis.
With the penetration of cloud computing and edge computing in the field of robotics, communication between multi-host systems, including within the edge and between the cloud and edges, has become increasingly important. In these cases, the network setups are mostly wireless connections, i.e. Wi-Fi or 4G, rather than Ethernet.

Although DDS addresses the need for real-time and reliable information dissemination, DDS was originally designed to be more suitable for ethernet environments, rather than wireless environments~\cite{kang2021comprehensive, peeroo2022exploring}. In an open or wireless environment, DDS discovery protocols could cause the flooding effect~\cite{liang2023performance}.
In the IoT, there are other networking middleware used for remote devices or machine-to-machine communication, such as Message Queuing Telemetry Transport (MQTT) and Zenoh.
MQTT~\cite{MQTT} is a lightweight messaging protocol that uses a publish/subscribe model for data transfer. Zenoh~\cite{zenoh} is an open-source protocol and suite of tools for data sharing and communication in distributed systems. Zenoh aims to provide a unified approach to data sharing and communication, regardless of the underlying hardware, network topology, or programming languages used.

Therefore, in Edge-to-Edge and Edge-to-Cloud communications based on ROS\,2  among multiple machines, it still remains a question of how the DDS combined with other networking middlewares can perform rather than DDS only.

\begin{figure}[h]
    \centering
    \includegraphics[width=0.72\textwidth, trim=150 0 150 0, clip]{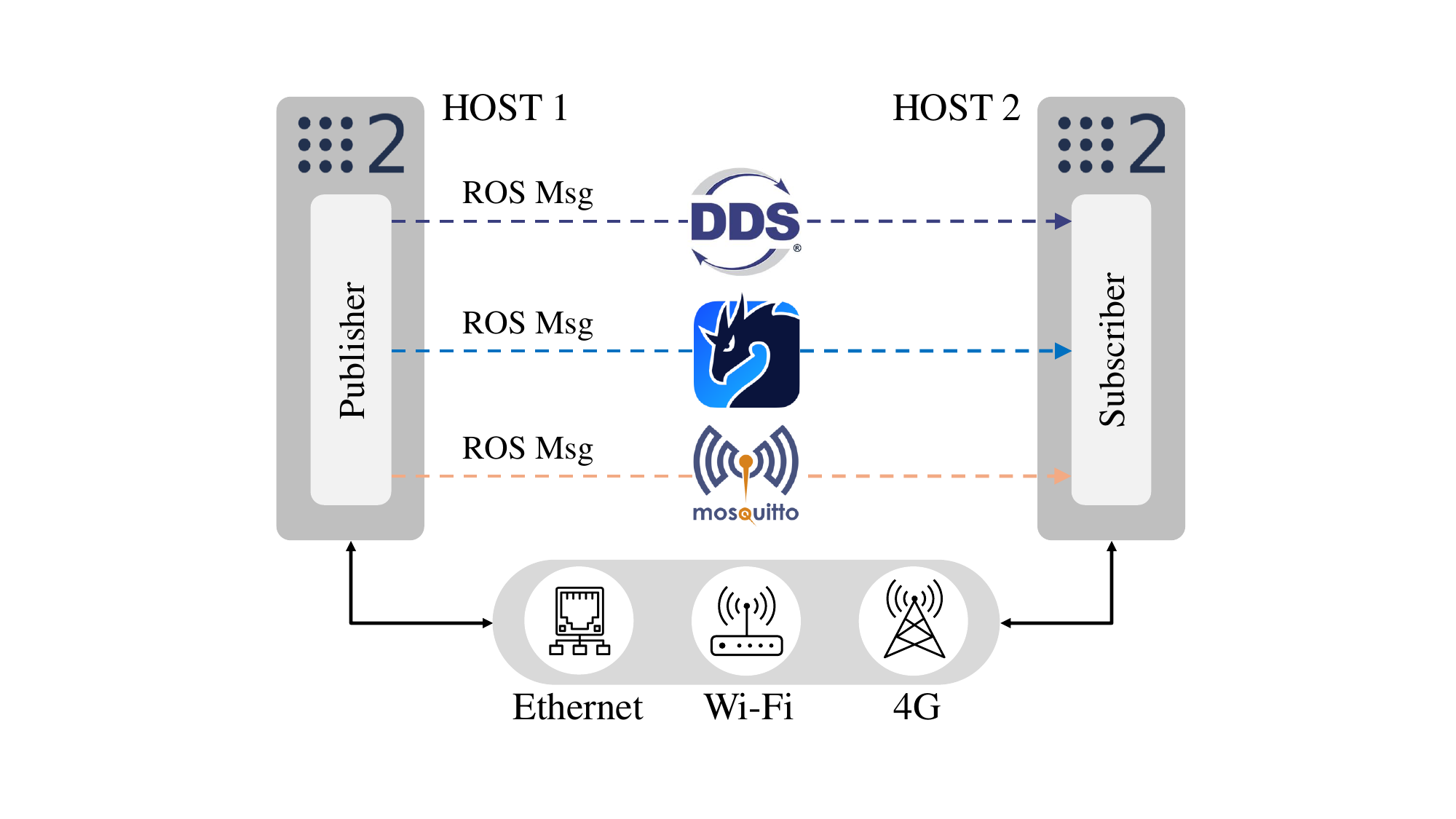}
    \caption{Proposed evaluation scheme to test DDS, Zenoh, and Mosquitto in different network setups}
    \label{fig:experiment_setup}
\end{figure}

To address the above issues, in this study, we provide a comprehensive evaluation of the communication performance among multi-host ROS\,2 systems, by solely relying on DDS and DDS integrated with MQTT or Zenoh. The quantitative analysis consists of the latency and throughput under various network setups including Ethernet, Wi-Fi, and 4G shown in Fig.~\ref{fig:experiment_setup}. Notably, the ROS messages utilized in the above experiment are of distinct types and sizes assuring a sufficient assessment. Additionally, by offering the drift error caused by the aforementioned communication approaches in the real-world moving experiment shown in Fig.~\ref{fig:realworld_setup}, we further extend our evaluation process.

To the best of our knowledge, extant literature lacks a comprehensive examination of how networking middleware affects the performance in a multi-host ROS\,2 system. 
Furthermore, the research landscape has not yet delved into the intricacies of assessing networking middleware performance within the context of tangible, real-world robotic systems. The present study, therefore, serves to bridge this evident gap in the scholarly discourse by conducting a meticulous comparative analysis of networking middleware functionalities utilizing ROS Messages and actualized experimentation on physical robotic platforms.

The rest of the paper is organized as follows. 
Section~\ref{sec:related_work} provides an overview of the background and related work, discussing the key concepts of the middleware used in ROS systems and presenting the existing literature on the performance study of middleware used in robotics and IoT. 
Section~\ref{sec:technology} describes the experimental setup and methodology used to evaluate the performance of these networking middlewares.  
Section~\ref{sec:results} presents the experiment results of DDS, MQTT, and Zenoh, including latency and throughput. We analyze the findings as well.
Section~\ref{sec:discussion} discusses CPU usage and the impact of the security feature on performance.
Finally, Section~\ref{sec:conclusion} concludes the paper.

%% file: sec/02_Background_and_Concepts.tex

\section{Background and Related Work}
\label{sec:related_work}
Through this section, we give a brief introduction to ROS\,2 and the most used communication protocols, as well as the existing study on the performance of middleware.

\subsection{ROS\,2 and SROS\,2}


Robot Operating System (ROS) has emerged as a popular middleware for developing robotic systems due to its modular architecture, open-source nature, and large community support. ROS provides a framework for communication between various modules of a robotic system using message passing and service calls. ROS\,2, the latest version of ROS, was released in 2015, with an emphasis on real-time systems, large-scale deployments, and distributed systems. ROS\,2 has improved modularity and supports different middleware for communication.

SROS\,2, short for Secure ROS\,2, is a security framework designed to enhance the data protection and communication security of robotic systems built on ROS\,2. SROS\,2 provides a set of tools, and libraries for providing security, authentication, and access control for ROS\,2 entities~\cite{mayoral2022sros2}.
In~\cite{kim2018security}, the authors investigate the trade-off between security and performance in ROS\,2 middleware. It provides guidelines for selecting appropriate security mechanisms and highlights the importance of a ``defense-in-depth'' approach. The authors conduct experiments to measure the impact of different security mechanisms on communication latency and throughput.



\subsection{Middleware}

\subsubsection{RMW}
ROS Middleware (RMW) is the abstraction layer responsible for facilitating communication between different components of a robotic system~\cite{maruyama2016exploring}. It provides a set of APIs and protocols for sending and receiving data between publishers and subscribers. The use of RMW allows for decoupling of the communication protocol from the application code, enabling interoperability between different components and platforms. 


\subsubsection{DDS}
\begin{figure}[t]
    \centering
    \includegraphics[width=0.88\textwidth, trim=200 10 200 10, clip]{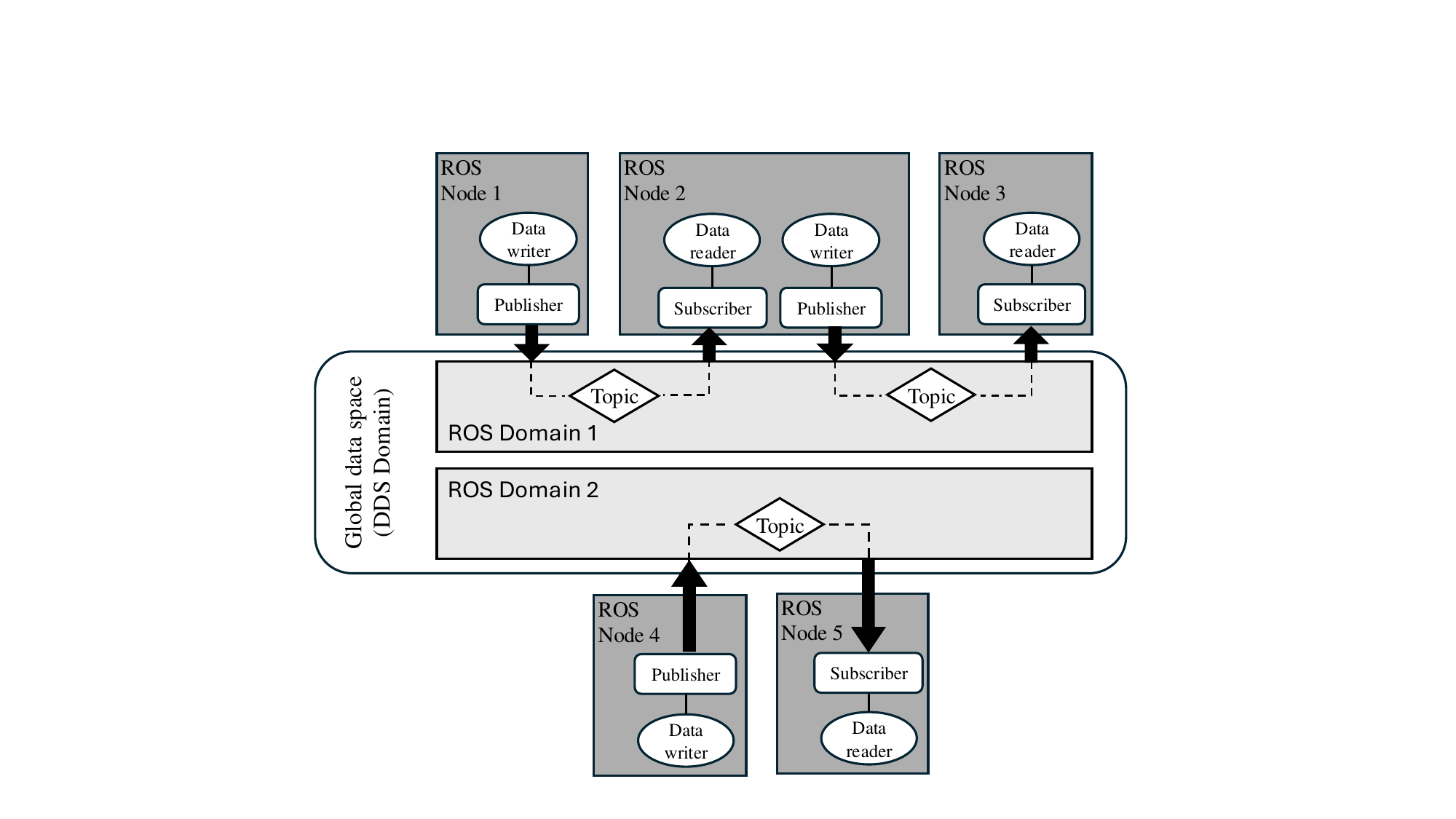}
    \caption{DDS architecture in the ROS\,2 system}
    \label{fig:dds_architecture}
\end{figure}

DDS is a publish-subscribe middleware that provides a data-centric communication model for distributed systems. It is designed for real-time systems and supports QoS settings to optimize communication performance. As the default RMW of ROS\,2, DDS provides a decentralized architecture shown in Fig.~\ref{fig:dds_architecture}, where nodes can communicate directly with each other without relying on a centralized broker. This architecture ensures reliability, fault tolerance, and scalability.

\subsubsection{MQTT}
MQTT is a lightweight publish-subscribe protocol designed for constrained devices and low-bandwidth network environments and has almost become a \textit{de-facto} standard in the field of IoT~\cite{yoshino2021highly, light2017mosquitto}. 
It supports QoS levels to ensure message delivery and allows for a wide range of applications, including sensors and IoT devices.


\subsubsection{Zenoh}
Zenoh is a new type of middleware that provides a unified data space for distributed systems. It is designed to provide seamless communication between edge devices, cloud systems, and hybrid environments. Zenoh provides a data-centric communication model and supports different data formats, including structured and unstructured data. It also provides a decentralized architecture and supports fault tolerance and scalability.

DDS, MQTT, and Zenoh are three different middleware options with distinct features. DDS is a data-centric middleware that offers a decentralized architecture, enabling direct peer-to-peer communication and extensive QoS settings for optimizing reliability, latency, and throughput. MQTT, on the other hand, is a lightweight publish-subscribe protocol designed for resource-constrained devices and networks, with a client-server architecture and low network overhead. It is suitable for IoT applications. Zenoh provides a unified data space, allowing seamless communication across edge devices, cloud systems, and hybrid environments. 

\subsection{Performance Tools}
In this part, we present the tools or packages that are developed to assess the performance of the ROS\,2 system.

\begin{itemize}
\item The performance\_test~\cite{performance_test}
tool tests latency and other performance metrics of various middleware implementations that support a pub/sub pattern. It is used to simulate the non-functional performance of the application.


\item The ros2\_tracing~\cite{bedard2022ros2tracing}
is a tool that provides tracing capabilities within ROS\,2. It enables the collection of detailed runtime data, including message passing, data flow, and timing characteristics. 

\item The ros-network-analysis~\cite{pandey2022empirical}
is a ROS package that provides tools to analyze the wireless network such as the signal quality, latency, throughput, link utilization, connection rates, error metrics, etc., between two ROS nodes, computers, or machines.

\item CARET (Chain-Aware ROS Evaluation Tool)~\cite{CARET} 
provides a customizable and analyzable framework for capturing and analyzing real-time execution traces in ROS-based systems. It allows developers to gain a deep understanding of the timing behavior and performance characteristics of their real-time applications, facilitating the development and optimization of reliable and efficient robotic systems.


\item The ros2\_latency\_evaluation~\cite{kronauer2021latency} provides tools and scripts for measuring and analyzing latency in ROS\,2 communication between multiple nodes. It includes code that sets up the necessary infrastructure to perform latency measurements, collects latency data during the execution of ROS\,2 systems, and provides analysis scripts to process and visualize the collected data.

\item The Autoware\_Perf~\cite{li2022autoware_perf} can measure the callback latency, node latency, and communication latency in ROS\,2 applications. In addition, Autoware\_Perf calculates the end-to-end latency by a convolutional integral as an estimated value.

\end{itemize}

\subsection{Related Work on the Performance of Middleware}
Several studies have investigated the performance and characteristics of different middleware solutions in the context of robotic systems and communication frameworks. These studies shed light on various aspects such as latency, throughput, queuing systems, dataflow architectures, and message analysis in distributed environments. 

Dhas Y J. et al.~\cite{dhas2019review} provided a qualitative and conceptual overview of IoT protocols and service-oriented middleware, offering insights into their roles, capabilities, and challenges in the Internet of Things domain.
Examining a decentralized serverless architecture,~\cite{escobar2023decentralized} addresses the challenges of dataflow in the Cloud-to-Edge continuum. The research presents an architectural solution that can provide insights into optimizing communication and data processing in the context of distributed systems. 
In~\cite{bedard2023message}, the authors focus on message flow analysis in distributed ROS\,2 systems, addressing complex causal relationships. The paper's insights into message propagation and interactions contribute to understanding the behavior of ROS Middleware in complex scenarios.
L. Puck et al. investigated the real-time performance of ROS\,2-based robotic control within a time-synchronized distributed network~\cite{puck2021performance}.
In~\cite{fu2020fair}, the authors use a standardized metric and reproducible experimental environment to provide a fair comparison among five message queuing systems, including Kafka, RabbitMQ, RocketMQ, ActiveMQ, and Pulsar.
The latency characteristics of multi-node systems within ROS\,2 are examined in~\cite{kronauer2021latency}. It highlights the criticality of low latency in robotics and presents an in-depth analysis of factors affecting latency in ROS\,2. The work sheds light on the significance of optimizing communication mechanisms for real-time applications.
A comprehensive performance study on Zenoh, MQTT, Kafka, and DDS is provided in~\cite{liang2023performance}. By evaluating throughput and latency, the study contributes to understanding the trade-offs and capabilities of different middleware implementations, thus aiding in informed middleware selection.


%% file: sec/03_Solutions.tex

\section{Experiment Methodology}
\label{sec:technology}
In this section, we outline the experimental methodology employed to evaluate and compare the performance of DDS, MQTT, and Zenoh in Edge-to-Edge and Edge-to-Cloud communication scenarios. We describe the experimental setup, test scenarios, the equipment utilized for testing, and performance metrics used for comparison.

\subsection{Experimental Setup}
Our experiments were conducted in a networked environment consisting of a multi-host ROS\,2 system.
As shown in Fig.\ref{fig:experiment_setup}, the setup includes a combination of devices
and communication channels, including Ethernet, Wi-Fi, and 4G connectivity. Each networking middleware (DDS, MQTT, and Zenoh) was configured and integrated into the ROS\,2 framework, enabling communication between the robotic systems. Notably, the RMW is DDS in all experiments. In the experiments of Zenoh and MQTT, we enabled ``Localhost Only'' and set bridge between DDS and Zenoh or MQTT, so that DDS will publish or subscribe ROS Messages only to or from the local host. In other words, for Zenoh or MQTT, the communication within one host used DDS, and communication between two hosts used Zenoh or MQTT.

Regarding the ROS\,2 DDS implementation, in this paper we use CycloneDDS due to its higher compatibility with Zenoh. It is out of the scope of this work to study the performance with different DDS implementations, since there are a number of studies and technical reports dedicated to this matter already in the literature~\cite{RMWreports, FastDDSPerformance, kronauer2021latency}. Owing to the fact that our work focuses on inter-host communication, we configure all hosts to use CycloneDDS as RMW for consistency across experiments.

\begin{figure}[t]
    \centering
    \includegraphics[width=0.35\textwidth]{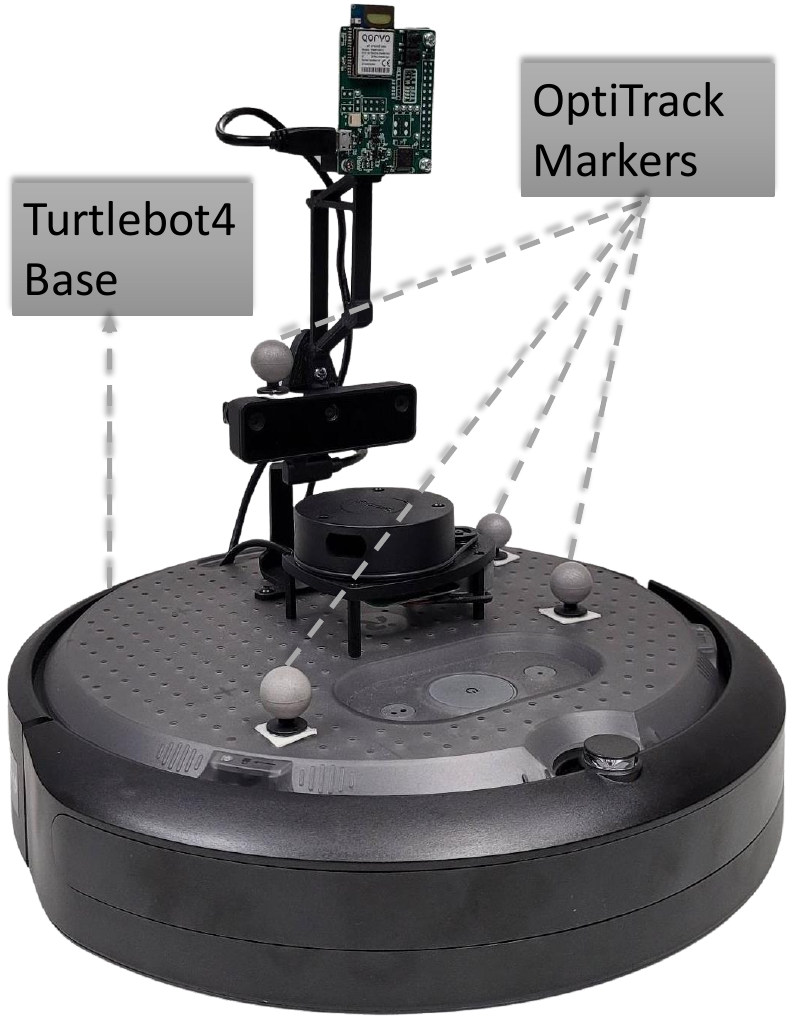}
    \caption{The Turtlebot 4 robot used in the experiment with OptiTrack markers on top}
    \label{fig:tb}
\end{figure}

\subsubsection{Hardware} \label{experiment_hardware}
The experiments were conducted in multi-machine scenarios. All devices involved in the experiments run Ubuntu 20.04. The device information is shown in Table.~\ref{tab:hardware}.

In order to evaluate the performance of different networking middleware implementations on a real-robot platform, we set the networking middlewares between a TurtleBot 4 robot and a laptop. The Turtlebot4 robot is shown in Fig.~\ref{fig:tb}. 

\begin{table}[h]
\centering
\caption{Hardware Setup}
\label{tab:hardware}
\begin{tabular}{@{}lccccl@{}}
\hline
\textbf{Devices} & \textbf{System}& \textbf{PROCESSOR} & \textbf{Memory} \\ \hline
Host 1         & Ubuntu 20.04           & AMD®Ryzen 7 5800h        & 16GB         \\ 
Host 2         & Ubuntu 20.04          & Intel® Core i3-1215U        & 16GB        \\ 
Router/Broker         & Ubuntu 20.04          & Intel® Core i7-9700K        & 64GB        \\ \hline
\end{tabular}
\end{table}

\begin{figure}[t]
    \centering
    \includegraphics[width=0.6\textwidth, trim=200 50 200 50, clip]{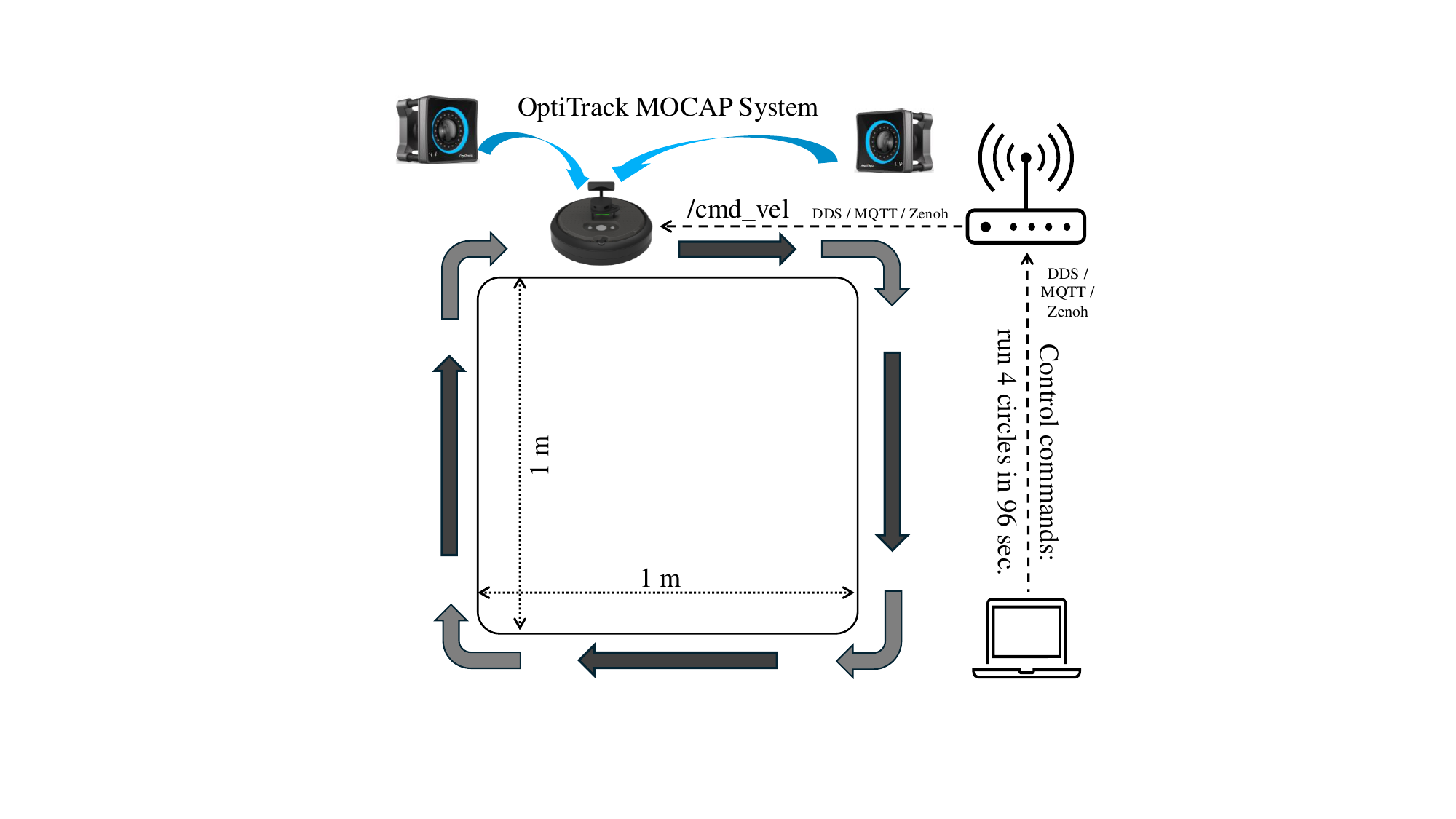}
    \caption{Actual robot setup. The TurtleBot 4 receives the ``/cmd\_vel'' velocity command from the HOST to run a square trajectory. The linear velocity is set to 0.33 m/s forwarding, and the turning velocity is set to 30 degree/s. The OptiTrack MOCAP System is only used to record the trajectories, which are subsequently employed in calculating the drift errors.}
    \label{fig:realworld_setup}
\end{figure}

As shown Fig.~\ref{fig:realworld_setup}, the TurtleBot 4 and the laptop both connect to the same Wi-Fi local network. The laptop sends velocity commands to the TurtleBot 4 to execute a square-shaped trajectory. The TurtleBot 4 will run four identical square circles in 96 seconds. The OptiTrack motion capture (MOCAP) system records the actual trajectory of the TurtleBot 4 during the execution of the square shape.

\subsubsection{Network}
\begin{figure*}[h]
	\centering
	\subfloat[Ethernet Local Network]{\includegraphics[width=.32\linewidth, trim=250 185 250 150, clip]{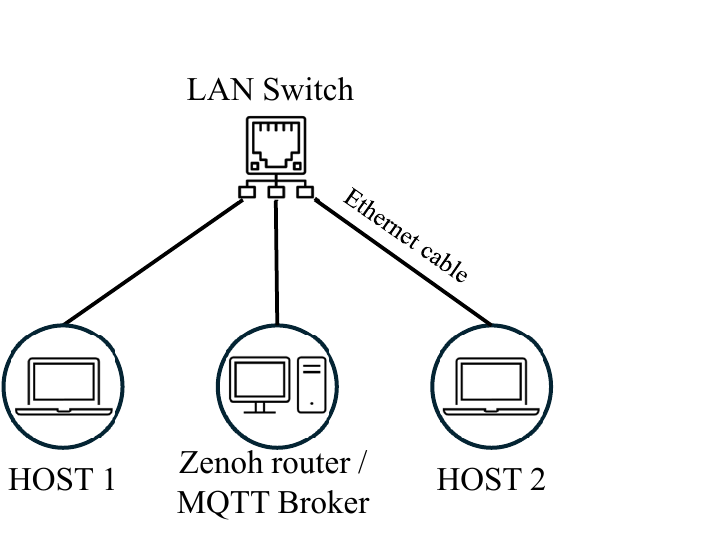}}\hspace{5pt}
	\subfloat[Wi-Fi Local Network]{\includegraphics[width=.32\linewidth]{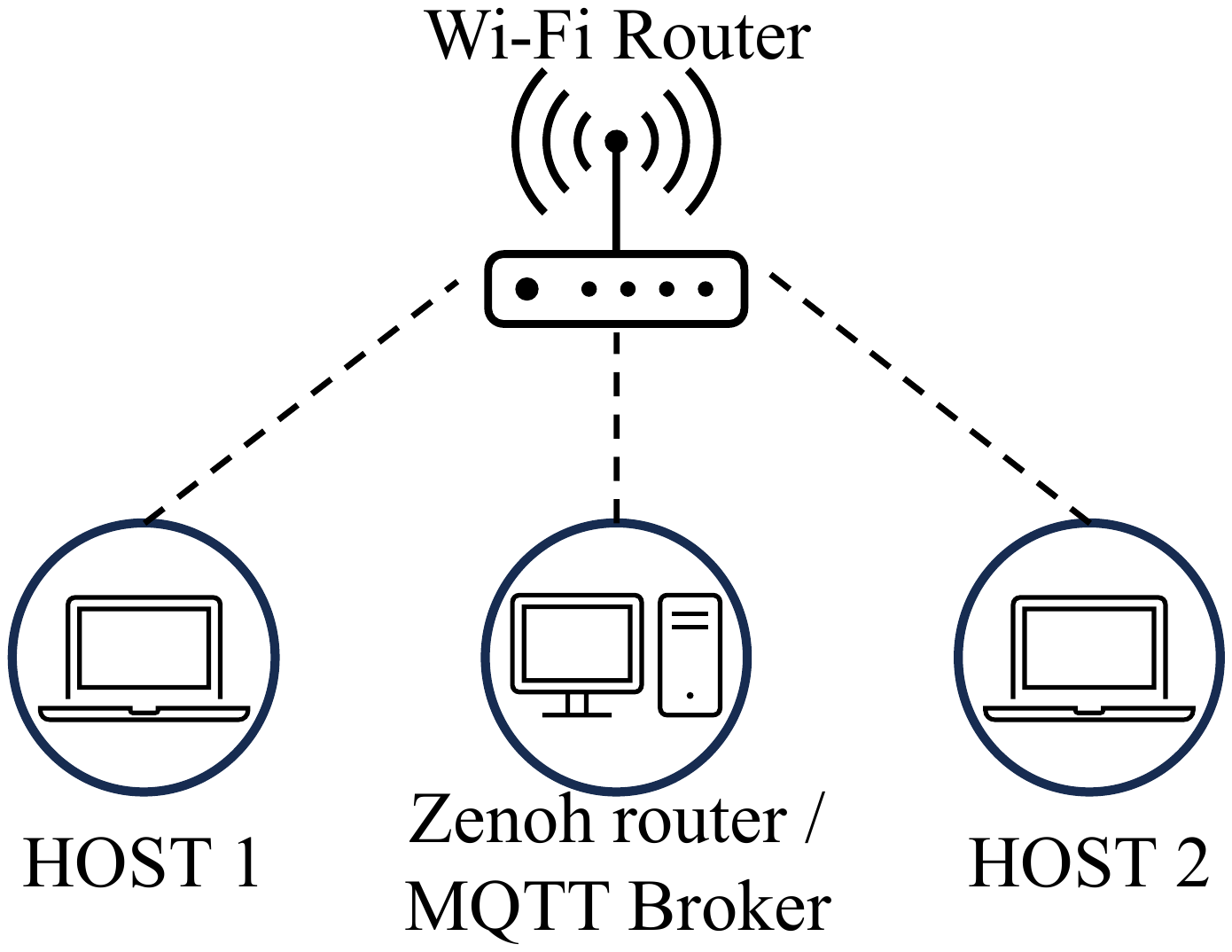}}
	\subfloat[4G (Zerotier Virtualized Network)]{\includegraphics[width=.32\linewidth]{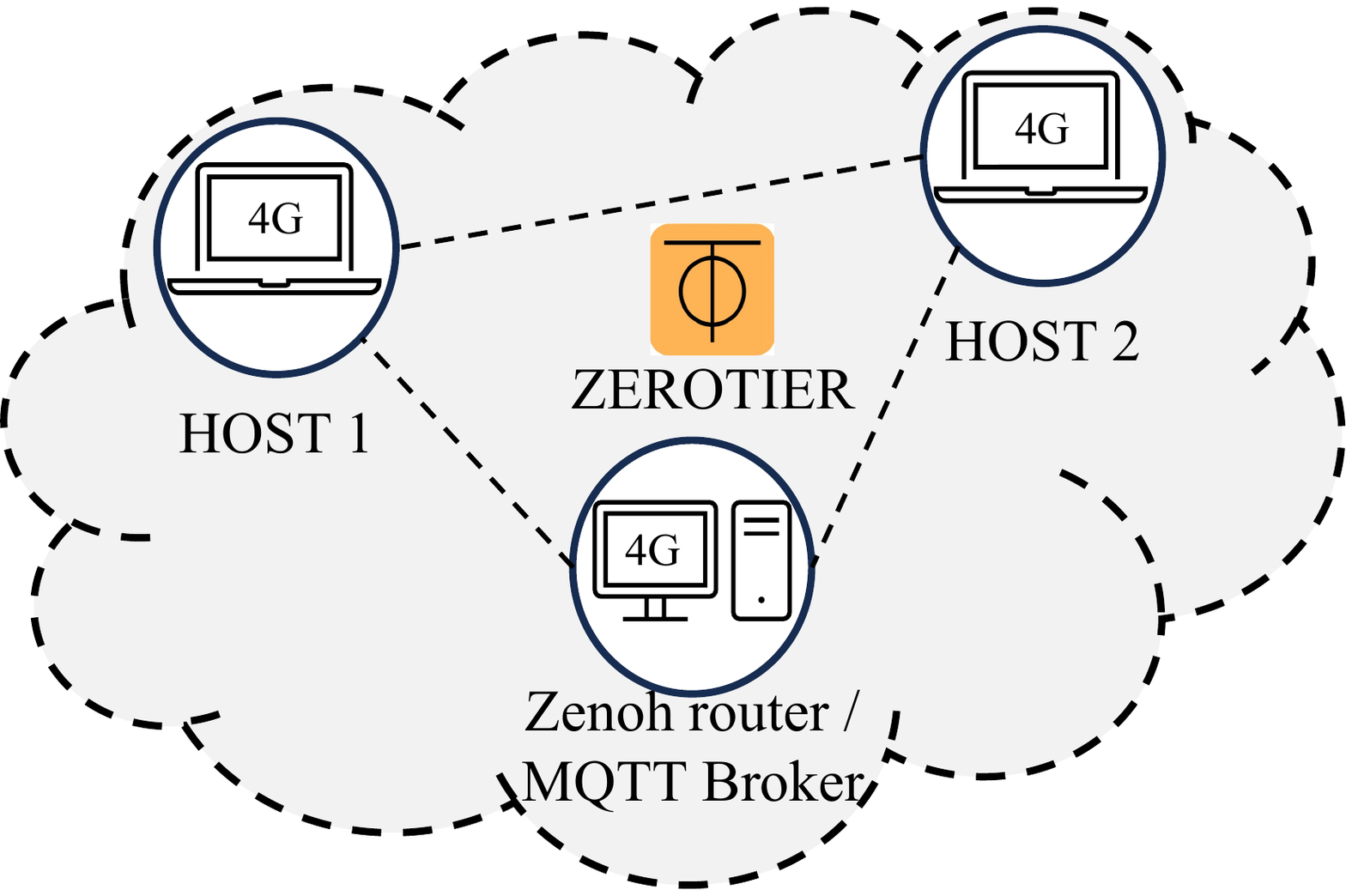}}\hspace{5pt}
	\caption{Network setups. The router or broker is for Zenoh and MQTT; DDS does not need an individual broker.}
    \label{fig:network_setups}
\end{figure*}

In order to assess the performance of the networking middleware in different network conditions, we set three different network environments, as shown in Fig.~\ref{fig:network_setups}.
The first setup utilized an Ethernet local network, representing a high-bandwidth and low-latency environment for Edge-to-Edge communication. The LAN Switch used in Ethernet Local Network is Unifi Flex Mini. The second setup involved a local network using Wi-Fi connectivity, simulating Edge-to-Edge communication within a confined area. The Wi-Fi Router is Huawei B593. Lastly, we leveraged the Zerotier virtualized network to 4G setup, which emulates Edge-to-Cloud communication scenarios. In this scenario, for HOST 1 \& 2, we used HUAWEI 4G Dongle E3372 as 4G module, and for broker, we used Netgear 4G LTE Mobile Router as 4G module.

\subsubsection{ROS Message}
To comprehensively evaluate the performance of the networking middleware, we employed various types and sizes of ROS Messages in our experiments, including \textit{Array1k},\textit{Array4k}, \textit{Array16k}, \textit{Array64k}, \textit{Array256k}, \textit{Array1m}, \textit{Array2m}, \textit{PointCloud512k}, \textit{PointCloud1m}, and \textit{PointCloud2m}, where \textit{k} and \textit{m} are \textit{Kilobyte} and \textit{Megabyte}. The ROS Messages were published and subscribed with a frequency of 10\,Hz. By using different sizes and types of ROS Messages, we aimed to assess the networking middleware's ability to handle messages of varying complexities and data sizes, simulating real-world scenarios encountered in robotic systems.


By designing our experimental setup to encompass different hardware configurations, network conditions, and message types, we ensured a comprehensive evaluation of the performance of DDS, MQTT, and Zenoh in Edge-to-Edge and Edge-to-Cloud communication scenarios. The chosen setups allowed us to capture the nuances and implications of each networking middleware's performance under various conditions, facilitating a robust comparison and analysis of their capabilities.

\subsection{Performance Metrics and Tool}
We focus on two key performance metrics including latency and throughput, following the methodology outlined by the DDS foundation, particularly for publish/subscribe protocols~\cite{xiong2010evaluating}. The performance tool selected for this research is performance\_test~\cite{performance_test}.
The primary objective of this work is to compare the performance of different networking middlewares in Edge-to-Edge and Edge-to-Cloud environments. The performance\_test tool is specifically designed for benchmarking and performance evaluation, making it well-suited for this purpose. It provides standardized methods for generating synthetic message traffic, measuring latency and throughput, and facilitating direct performance comparisons between different middlewares. In addition, the performance\_test supports SROS\,2 with a simple setup.

The use of the performance\_test, enabled accurate measurement of latency and throughput. To ensure reliable data collection, we configured the QoS reliability setting to be reliable, guaranteeing message delivery without loss. The reliable QoS settings ensured that our data collection was robust, minimizing the impact of data loss or distortion.

%% file: sec/04_Applications.tex
\section{Experiment Results}
\label{sec:results}
In this section, we present the results of our experiments comparing the performance of DDS, MQTT, and Zenoh in different network conditions and on an actual robot platform.
In the subsection of{~\ref{latency_result}} Latency and{~\ref{throughput_result}} Throughput, we present five cases, including CycloneDDS, Zenoh-TCP, Zenoh-broker, MQTT-no-broker, MQTT-broker. CycloneDDS is the default RMW implementation in ROS\,2. In this configuration, the middleware is CycloneDDS alone. In the Zenoh-TCP configuration, TCP is used as the transport protocol for peer-to-peer communication. MQTT is a lightweight messaging protocol using a publish/subscribe model. In the MQTT-no-broker configuration, one host is set as the role of broker, and the communication occurs directly between two hosts without a centralized broker. In the Zenoh-broker or MQTT-broker configuration, a broker-based architecture is set to manage communication between hosts.

\subsection{Latency} \label{latency_result}
Table~\ref{tab:latency} presents the mean latency results obtained from the experiments conducted in different network setups, including Ethernet, Wi-Fi, and 4G. In each network setup, we measured the mean latency for different middleware implementations and various ROS message types. The results provide insights into the latency performance of each middleware under different network conditions and message sizes.

\begin{table*}[t]
    \centering
     \caption{Mean latency with different network setup}
     \label{tab:latency}

    \begin{subtable}{0.98\textwidth}
        \centering
        \caption{Mean latency (ms) with Ethernet}
        \label{tab:latency_ethernet}
\resizebox{0.96\textwidth}{!}{
        \begin{tabular}{@{}lccccccl@{}}
        \toprule

        \textbf{ROS Message} & \textbf{CycloneDDS} & \textbf{Zenoh-TCP} & \textbf{Zenoh-broker} & \textbf{MQTT-nobroker} & \textbf{MQTT-broker} \\ 
        \midrule
        Array1k      & \textbf{1.29}	 & 1.98	 & 1.78 & 	89.74 & 	91.01 \\ 
        Array4k        & 	\textbf{1.30} & 	1.97	 & 1.91	 & 86.59	 & 3.87       \\ 
        Array16k   & \textbf{1.55} & 	2.36 & 	2.24 & 	86.98 & 	3.58       \\ 
        Array64k        & 	3.35 & 	3.35	 & \textbf{3.27} & 	3.51 & 	14.77       \\ 
        Array256k    & 	\textbf{5.37} & 	7.42 & 	7.42	 & 10.75	 & 48.50     \\ 
        Array1m      & \textbf{19.28} & 	32.88	 & 42.69 & 	136.60	 & 5545.04     \\ 
        Array2m     & 	\textbf{37.97} & 	70.97 & 	70.61 & 	312.89 & 	7468.62     \\ 
        PointCloud512k    & 	\textbf{11.16} & 	20.51 & 	18.47 & 	18.45	 & 87.07       \\ 
        PointCloud1m    & 	\textbf{21.95} & 	37.42 & 	41.24 & 	94.92	 & 1020.70       \\ 
        PointCloud2m     & 	\textbf{39.94} & 	76.78 & 	77.12 & 	540.25	 & 5625.40       \\
        \bottomrule
        \end{tabular}
}
     \end{subtable}
    \hfill
    \begin{subtable}{0.98\textwidth}
        \centering
        \label{tab:latency_Wi-Fi}
        \caption{Mean latency with Wi-Fi}
\resizebox{0.96\textwidth}{!}{
        \begin{tabular}{@{}lccccccl@{}}
        \toprule

        \textbf{ROS Message} & \textbf{CycloneDDS} & \textbf{Zenoh-TCP} & \textbf{Zenoh-broker} & \textbf{MQTT-nobroker} & \textbf{MQTT-broker} \\ 
        \midrule
        Array1k         & 	51.40 & 	14.82 & 	\textbf{12.58} & 	92.93	 & 173.62\\ 
        Array4k       	 & 39.84	 & \textbf{32.34} & 	54.11 & 	101.19 & 	73.64   \\ 
        Array16k        & 	56.73	 & \textbf{48.30}	 & 139.63 & 	101.77 & 	125.69    \\ 
        
        Array64k         	 & 606.50 & 	452.51 & 	326.96	 & \textbf{278.81} & 	4382.25  \\ 
        Array256k   	 & 1574.14 & 	3021.77	 & 2323.91	 & 13538.75	 & 19788.57 \\ 
        Array1m   	 & - & 	\textbf{5960.16}	 & 7599.04	 & 20950.00	 & -  \\ 
        Array2m    & - & 	\textbf{9386.79} & 	11337.42 & 	- & 	-   \\ 
        PointCloud512k    & 	-	 & \textbf{5164.90} & 	5407.28	 & 11307.95	 & -    \\ 
        PointCloud1m    & 	\textbf{3748.00}	 & 5326.69 & 	5657.96	 & 24115.00	 & -    \\ 
        PointCloud2m   	 & - & 	9283.26	 & \textbf{8805.96} & 	- & 	-     \\ 
        \bottomrule
        \end{tabular}
}
     \end{subtable}
    \hfill
    \begin{subtable}{0.98\textwidth}
        \centering
        \caption{Mean latency with 4G}
        \label{tab:latency_4G}
\resizebox{0.96\textwidth}{!}{
        \begin{tabular}{@{}lccccccl@{}}
        \toprule

        \textbf{ROS Message} & \textbf{CycloneDDS} & \textbf{Zenoh-TCP} & \textbf{Zenoh-broker} & \textbf{MQTT-nobroker} & \textbf{MQTT-broker} \\
        \midrule
        Array1k        	 & \textbf{58.80} & 	72.11	 & 134.02 & 	279.17 & 	679.91\\ 
        Array4k       & 	103.75 & 	\textbf{102.86} & 	98.05 & 	228.84	 & 537.87\\ 
        Array16k        & 	\textbf{116.95}	 & 150.81 & 	145.42 & 	220.82 & 	530.20\\ 
        Array64k         & \textbf{163.29} & 	3349.90 & 	434.19 & 	541.76 & 	2222.11\\ 
        Array256k   & -	 & \textbf{7458.61}	 & 7556.25	 & 10736.73	 & 19987.27\\ 
        Array1m      & -	 & \textbf{10086.88}	 & 11811.52	 & -	 & - \\ 
        Array2m    	 & -	 & 15365.29	 & \textbf{10870.59}	 & -	 & -\\ 
        PointCloud512k    	 & -	 & 7357.50	 & \textbf{6591.85}	 & -	 & - \\ 
        PointCloud1m   	 & -	 & 13254.00	 & \textbf{10333.13}	 & -	 & - \\ 
        PointCloud2m   	 & -	 & 14309.50 & 	\textbf{11900.27}	 & -	 & -  \\ 
        
        \bottomrule
        
        \end{tabular}
}
     \end{subtable}

\end{table*}

It is evident that the mean latency increases as the message size grows in all network scenarios.
The results also demonstrate the impact of different network setups on the latency performance of networking middleware. Each middleware demonstrates superior performance under Ethernet conditions, which offer the highest bandwidth, compared to Wi-Fi and 4G environments. Notably, CycloneDDS performs best in Ethernet setup, while Zenoh performs best in Wi-Fi and 4G setups. This is caused by the DDS discovery mechanism. DDS leverages UDP multicast features to broadcast messages in the transport layer. It enables fast peer discovery and QoS-based low-latency data transmission in wired local networks where bandwidth and packet loss rate are promised. On the other hand, in wireless networks such as Wi-Fi or 4G, UDP multicast could cause the flooding effect, resulting in poorer performance~\cite{liang2023performance, ddsoverhead, peeroo2022exploring}. 
These findings highlight the importance of considering network conditions and message characteristics when selecting an appropriate middleware for Edge-to-Edge and Edge-to-Cloud communication in ROS-based robotic systems.



\subsection{Throughput} \label{throughput_result}
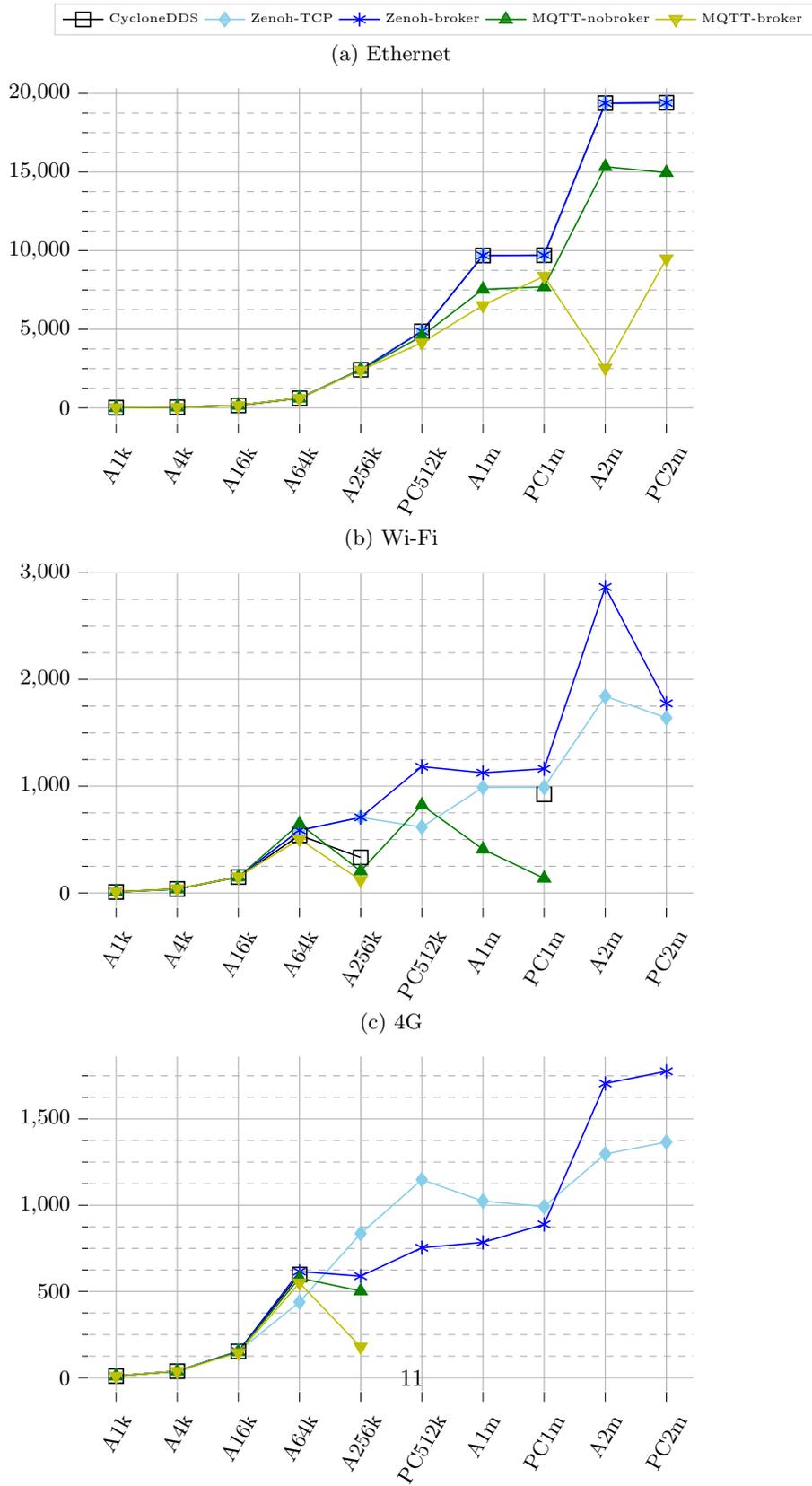
\begin{figure}[H]
	\centering
        \scriptsize{\input{fig/throughput.tex}}
	\caption{Throughput (Kilobytes/sec) with different network setup. In Subfig (a), since the data difference is very small, the lines of CycloneDDS, Zenoh-TCP, and Zenoh-broker overlap. }
    \label{fig:throughput}
\end{figure}

Fig.~\ref{fig:throughput} presents the throughput results. 
The results showed that the bandwidth of the network has the largest impact on throughput. 
Furthermore, as illustrated in Fig.{~\ref{fig:throughput}}(a), CycloneDDS exhibited comparable performance to Zenoh under Ethernet conditions; however, Zenoh outperformed CycloneDDS in Wi-Fi and 4G scenarios. 

\subsection{Performance on Actual Robot Platform}
\begin{figure}[h]
    \centering
    \scriptsize{\input{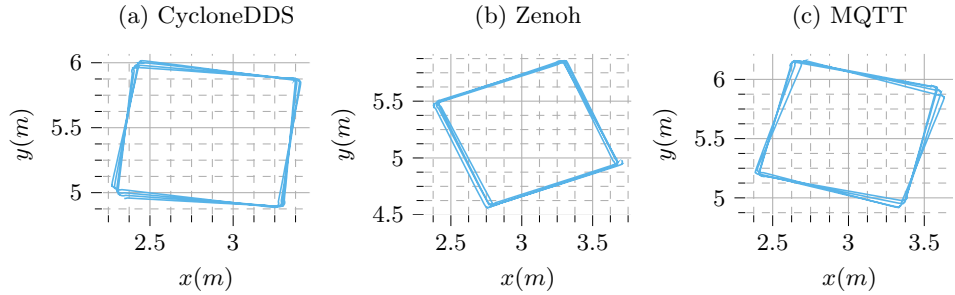}}
    \caption{Robot moving trajectory over 96\,s of different middleware. The trajectories are for illustration and are not comparable.}
    \label{fig:trajectory_real}
\end{figure}

\begin{figure}[H]
    \centering
    \setlength{\figurewidth}{0.7\textwidth}
    \setlength{\figureheight}{0.5\textwidth}
    \scriptsize{\input{fig/ex_plot}}
    \caption{Robot drift over 96 seconds for various middlewares, analyzed using data from 2200 frames per middleware.}
    \label{fig:trajectory_boxplot}
\end{figure}
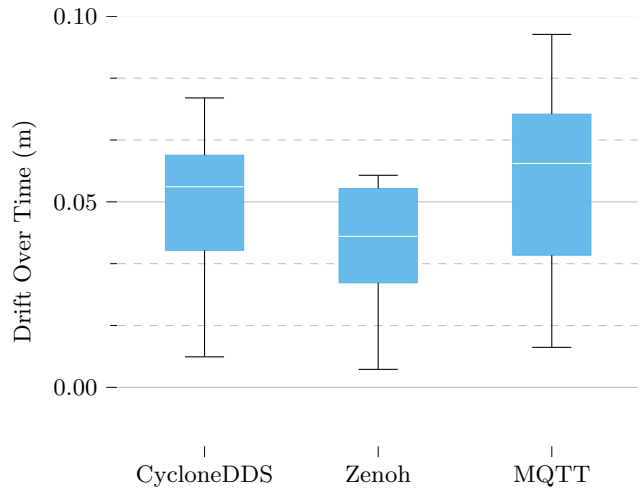

As mentioned in{~\ref{experiment_hardware}}, we conducted an experiment on a real-robot platform, TurtleBot 4. The setup is shown in Fig.{~\ref{fig:realworld_setup}}. The TurtleBot 4 and the laptop, connected to the same Wi-Fi network, enable the laptop to send velocity commands for the TurtleBot 4 to execute a square-shaped trajectory. At the same time, the OptiTrack motion capture system records the actual trajectory. Fig.~\ref{fig:trajectory_real} shows the moving trajectory of the TurtleBot 4 over 96\,s. Fig.~\ref{fig:trajectory_boxplot} shows the drift error between the last and the first circle.

Zenoh demonstrated the closest adherence to the square trajectory, exhibiting minimal drift and accurately reproducing the desired path. This result highlights Zenoh's robustness and capability to maintain precise position and movement control on a real robot platform. On the other hand, the trajectories of CycloneDDS and MQTT showed varying degrees of drift over time.

\begin{table*}[t]
\centering
 \caption{CPU usage (\%) with different network setup. The data is in the format of (Ethernet\_CPU, Wi-Fi\_CPU, 4G\_CPU).}
 \label{tab:cpu}
\begin{adjustbox}{width=0.98\textwidth}

\begin{tabular}{@{}lccccccl@{}}
\toprule
\textbf{MSG}& \textbf{CycloneDDS} & \textbf{Zenoh-TCP} & \textbf{Zenoh-broker} & \textbf{MQTT-nobroker} & \textbf{MQTT-broker} \\ 
\midrule
Array1k  & (0.04, 0.04, 0.04) & (0.05, 0.05, 0.06) & (0.06, 0.05, 0.06) & (0.05, 0.05, 0.05) & (0.05, 0.05, 0.05) \\ 
Array4k  & (0.04, 0.03, 0.04) & (0.05, 0.05, 0.06) & (0.06, 0.05, 0.06) & (0.05, 0.05, 0.05) & (0.05, 0.05, 0.05) \\ 
Array16k  & (0.05, 0.04, 0.05) & (0.06, 0.06, 0.06) & (0.06, 0.06, 0.06) & (0.05, 0.05, 0.05) & (0.04, 0.05, 0.06) \\ 
Array64k  & (0.07, 0.06, 0.06) & (0.06, 0.05, 0.05) & (0.07, 0.05, 0.06) & (0.05, 0.05, 0.06) & (0.05, 0.03, 0.06) \\ 
Array256k  & (0.19, 0.09, 0.06) & (0.09, 0.05, 0.04) & (0.09, 0.05, 0.05) & (0.08, 0.05, 0.05) & (0.07, 0.04, 0.05) \\ 
Array1m  & (0.51, 0.08, 0.11) & (0.16, 0.04, 0.03) & (0.18, 0.03, 0.04) & (0.18, 0.13, 0.13) & (0.17, 0.12, 0.14) \\ 
Array2m & (0.77, 0.08, 0.19) & (0.30, 0.04, 0.03) & (0.32, 0.04, 0.03) & (0.30, 0.17, 0.19) & (0.20, 0.18, 0.19) \\ 
PointCloud512k  & (0.30, 0.07, 0.07) & (0.12, 0.06, 0.03) & (0.12, 0.04, 0.04) & (0.12, 0.08, 0.07) & (0.12, 0.09, 0.09) \\ 
PointCloud1m  & (0.57, 0.09, 0.12) & (0.18, 0.04, 0.03) & (0.19, 0.04, 0.04) & (0.17, 0.12, 0.14) & (0.21, 0.15, 0.14) \\ 
PointCloud2m & (0.84, 0.11, 0.19) & (0.32, 0.03, 0.03) & (0.30, 0.03, 0.04) & (0.33, 0.18, 0.20) & (0.22, 0.22, 0.20) \\ 

\bottomrule  
\end{tabular}
\end{adjustbox}
\end{table*}

%% file: fig/throughput.tex
\begin{tikzpicture}

\definecolor{darkgray176}{RGB}{176,176,176}
\definecolor{goldenrod1911910}{RGB}{191,191,0}
\definecolor{green01270}{RGB}{0,127,0}
\definecolor{lightgray204}{RGB}{204,204,204}
\definecolor{skyblue}{RGB}{135,206,235}

\begin{groupplot}[group style={group size=1 by 3,vertical sep=2.2cm},
width = 0.8\textwidth, height=0.5\textwidth
]
\nextgroupplot[
axis line style={white},
tick align=outside,
tick pos=left,
title={(a) Ethernet},
unbounded coords=jump,
x grid style={darkgray176},
xmajorgrids,
xmin=-0.45, xmax=9.45,
xtick style={color=black},
xticklabel style={rotate=60},
xtick={0,1,2,3,4,5,6,7,8,9},
xtick={0,1,2,3,4,5,6,7,8,9},
xtick={0,1,2,3,4,5,6,7,8,9},
xtick={0,1,2,3,4,5,6,7,8,9},
xtick={0,1,2,3,4,5,6,7,8,9},
xtick={0,1,2,3,4,5,6,7,8,9},
xtick={0,1,2,3,4,5,6,7,8,9},
xticklabels={A1k,A4k,A16k,A64k,A256k,PC512k,A1m,PC1m,A2m,PC2m},
xticklabels={A1k,A4k,A16k,A64k,A256k,PC512k,A1m,PC1m,A2m,PC2m},
xticklabels={A1k,A4k,A16k,A64k,A256k,PC512k,A1m,PC1m,A2m,PC2m},
xticklabels={A1k,A4k,A16k,A64k,A256k,PC512k,A1m,PC1m,A2m,PC2m},
xticklabels={A1k,A4k,A16k,A64k,A256k,PC512k,A1m,PC1m,A2m,PC2m},
xticklabels={A1k,A4k,A16k,A64k,A256k,PC512k,A1m,PC1m,A2m,PC2m},
xticklabels={A1k,A4k,A16k,A64k,A256k,PC512k,A1m,PC1m,A2m,PC2m},
y grid style={darkgray176},
ymajorgrids,
ymin=-959.75, ymax=20374.75,
scaled ticks=false,
ytick style={color=black},
ymajorgrids,
ymajorticks=true,
minor y tick num = 3,
minor y grid style={dashed},
yminorgrids,
]
\addplot [semithick, black, mark=square, mark size=3, mark options={solid}]
table {%
0 10
1 38
2 151
3 606
4 2423
5 4868
6 9692
7 9712
8 19382
9 19405
};
\addplot [semithick, skyblue, mark=diamond*, mark size=3, mark options={solid}]
table {%
0 10
1 38
2 151
3 606
4 2423
5 4867
6 9691
7 9713
8 19381
9 19404
};
\addplot [semithick, blue, mark=asterisk, mark size=3, mark options={solid}]
table {%
0 10
1 38
2 152
3 606
4 2423
5 4868
6 9691
7 9713
8 19381
9 19404
};
\addplot [semithick, green01270, mark=triangle*, mark size=3, mark options={solid}]
table {%
0 10
1 38
2 151
3 603
4 2423
5 4559
6 7541
7 7695
8 15338
9 14963
};
\addplot [semithick, goldenrod1911910, mark=triangle*, mark size=3, mark options={solid,rotate=180}]
table {%
0 10
1 38
2 150
3 606
4 2380
5 4148
6 6518
7 8379
8 2525
9 9488
};
\nextgroupplot[
legend cell align={left},
legend columns=6,
legend style={
  fill opacity=0.8,
  draw opacity=1,
  text opacity=1,
  at={(1.1,2.6)},
  anchor=south east,
  draw=lightgray204,
  font=\tiny,
},
axis line style={white},
tick align=outside,
tick pos=left,
title={(b) Wi-Fi},
unbounded coords=jump,
x grid style={darkgray176},
xmajorgrids,
xmin=-0.45, xmax=9.45,
xtick style={color=black},
xticklabel style={rotate=60},
xtick={0,1,2,3,4,5,6,7,8,9},
xtick={0,1,2,3,4,5,6,7,8,9},
xtick={0,1,2,3,4,5,6,7,8,9},
xtick={0,1,2,3,4,5,6,7,8,9},
xtick={0,1,2,3,4,5,6,7,8,9},
xtick={0,1,2,3,4,5,6,7,8,9},
xtick={0,1,2,3,4,5,6,7,8,9},
xticklabels={A1k,A4k,A16k,A64k,A256k,PC512k,A1m,PC1m,A2m,PC2m},
xticklabels={A1k,A4k,A16k,A64k,A256k,PC512k,A1m,PC1m,A2m,PC2m},
xticklabels={A1k,A4k,A16k,A64k,A256k,PC512k,A1m,PC1m,A2m,PC2m},
xticklabels={A1k,A4k,A16k,A64k,A256k,PC512k,A1m,PC1m,A2m,PC2m},
xticklabels={A1k,A4k,A16k,A64k,A256k,PC512k,A1m,PC1m,A2m,PC2m},
xticklabels={A1k,A4k,A16k,A64k,A256k,PC512k,A1m,PC1m,A2m,PC2m},
xticklabels={A1k,A4k,A16k,A64k,A256k,PC512k,A1m,PC1m,A2m,PC2m},
y grid style={darkgray176},
ymajorgrids,
ymin=-133.85, ymax=3008.85,
ytick style={color=black},
ymajorgrids,
ymajorticks=true,
minor y tick num = 3,
minor y grid style={dashed},
yminorgrids,
]
\addplot [semithick, black, mark=square, mark size=3, mark options={solid}]
table {%
0 9
1 37
2 149
3 540
4 333
5 nan
6 nan
7 923
8 nan
9 nan
};
\addlegendentry{CycloneDDS}
\addplot [semithick, skyblue, mark=diamond*, mark size=3, mark options={solid}]
table {%
0 10
1 38
2 150
3 582
4 708
5 617
6 989
7 
8 1842
9 1640
};
\addlegendentry{Zenoh-TCP}
\addplot [semithick, blue, mark=asterisk, mark size=3, mark options={solid}]
table {%
0 10
1 38
2 152
3 587
4 708
5 1183
6 1126
7 1163
8 2866
9 1776
};
\addlegendentry{Zenoh-broker}
\addplot [semithick, green01270, mark=triangle*, mark size=3, mark options={solid}]
table {%
0 9
1 37
2 148
3 648
4 205
5 823
6 409
7 137
8 nan
9 nan
};
\addlegendentry{MQTT-nobroker}
\addplot [semithick, goldenrod1911910, mark=triangle*, mark size=3, mark options={solid,rotate=180}]
table {%
0 10
1 40
2 151
3 503
4 119
5 nan
6 nan
7 nan
8 nan
9 nan
};
\addlegendentry{MQTT-broker}

\nextgroupplot[
axis line style={white},
tick align=outside,
tick pos=left,
title={(c) 4G},
unbounded coords=jump,
x grid style={darkgray176},
xmajorgrids,
xmin=-0.45, xmax=9.45,
xtick style={color=black},
xticklabel style={rotate=60},
xtick={0,1,2,3,4,5,6,7,8,9},
xtick={0,1,2,3,4,5,6,7,8,9},
xtick={0,1,2,3,4,5,6,7,8,9},
xtick={0,1,2,3,4,5,6,7,8,9},
xtick={0,1,2,3,4,5,6,7,8,9},
xtick={0,1,2,3,4,5,6,7,8,9},
xtick={0,1,2,3,4,5,6,7,8,9},
xticklabels={A1k,A4k,A16k,A64k,A256k,PC512k,A1m,PC1m,A2m,PC2m},
xticklabels={A1k,A4k,A16k,A64k,A256k,PC512k,A1m,PC1m,A2m,PC2m},
xticklabels={A1k,A4k,A16k,A64k,A256k,PC512k,A1m,PC1m,A2m,PC2m},
xticklabels={A1k,A4k,A16k,A64k,A256k,PC512k,A1m,PC1m,A2m,PC2m},
xticklabels={A1k,A4k,A16k,A64k,A256k,PC512k,A1m,PC1m,A2m,PC2m},
xticklabels={A1k,A4k,A16k,A64k,A256k,PC512k,A1m,PC1m,A2m,PC2m},
xticklabels={A1k,A4k,A16k,A64k,A256k,PC512k,A1m,PC1m,A2m,PC2m},
y grid style={darkgray176},
ymajorgrids,
ymin=-79.35, ymax=1864.35,
ytick style={color=black},
ymajorgrids,
ymajorticks=true,
minor y tick num = 3,
minor y grid style={dashed},
yminorgrids,
]

\addplot [semithick, black, mark=square, mark size=3, mark options={solid}]
table {%
0 10
1 38
2 153
3 599
4 nan
5 nan
6 nan
7 nan
8 nan
9 nan
};
\addplot [semithick, skyblue, mark=diamond*, mark size=3, mark options={solid}]
table {%
0 10
1 38
2 154
3 439
4 836
5 1148
6 1024
7 992
8 1297
9 1366
};
\addplot [semithick, blue, mark=asterisk, mark size=3, mark options={solid}]
table {%
0 10
1 38
2 153
3 616
4 589
5 754
6 785
7 889
8 1706
9 1776
};
\addplot [semithick, green01270, mark=triangle*, mark size=3, mark options={solid}]
table {%
0 10
1 38
2 151
3 578
4 503
5 nan
6 nan
7 nan
8 nan
9 nan
};
\addplot [semithick, goldenrod1911910, mark=triangle*, mark size=3, mark options={solid,rotate=180}]
table {%
0 9
1 38
2 145
3 550
4 179
5 nan
6 nan
7 nan
8 nan
9 nan
};
\end{groupplot}

\end{tikzpicture}

%% file: fig/ex_plot.tex
\begin{tikzpicture}


\definecolor{color0}{rgb}{0.90, 0.62, 0.00}
\definecolor{color1}{rgb}{0.34, 0.70, 0.91}
\definecolor{color2}{rgb}{0.00, 0.62, 0.45}
\definecolor{color3}{rgb}{0.94, 0.89, 0.27}
\definecolor{color4}{rgb}{0.00, 0.45, 0.69}
\definecolor{color5}{rgb}{0.83, 0.37, 0.00}

\begin{axis}[
    width=\figurewidth,
    height=\figureheight,
axis line style={white},
tick align=outside,
tick pos=left,
xmin=1.5, xmax=4.5,
xtick style={color=black},
xtick={2,3,4},
xticklabels={CycloneDDS, Zenoh, MQTT},
ymajorgrids,
ymajorticks=true,
minor y tick num = 2,
minor y grid style={dashed},
yminorgrids,
ylabel={Drift Over Time (m)},
ymin=-0.015744177416906, ymax=0.10,
ytick style={color=black},
ytick={-0.05,0,0.05,0.1,0.15,0.2,0.25,0.3,0.35},
yticklabels={\ensuremath{-}0.05,0.00,0.05,0.10,0.15,0.20,0.25,0.30,0.35}
]

\addplot [color1, fill=color1, fill opacity=0.9]
table {%
1.775 0.0369475758899489
2.225 0.0369475758899489
2.225 0.0625648292325726
1.775 0.0625648292325726
1.775 0.0369475758899489
};
\addplot [black]
table {%
2 0.0369475758899489
2 0.00822611335929743
};
\addplot [black]
table {%
2 0.0625648292325726
2 0.0780078630395681
};
\addplot [black]
table {%
1.8875 0.00822611335929743
2.1125 0.00822611335929743
};
\addplot [black]
table {%
1.8875 0.0780078630395681
2.1125 0.0780078630395681
};
\addplot [color1, fill=color1, fill opacity=0.9]
table {%
2.775 0.0281793214973548
3.225 0.0281793214973548
3.225 0.0535809724272519
2.775 0.0535809724272519
2.775 0.0281793214973548
};
\addplot [black]
table {%
3 0.0281793214973548
3 0.0048473692865307
};
\addplot [black]
table {%
3 0.0535809724272519
3 0.0571603537515299
};
\addplot [black]
table {%
2.8875 0.0048473692865307
3.1125 0.0048473692865307
};
\addplot [black]
table {%
2.8875 0.0571603537515299
3.1125 0.0571603537515299
};
\addplot [color1, fill=color1, fill opacity=0.9]
table {%
3.775 0.0356275788259603
4.225 0.0356275788259603
4.225 0.0736173214671656
3.775 0.0736173214671656
3.775 0.0356275788259603
};
\addplot [black]
table {%
4 0.0356275788259603
4 0.0107763558311707
};
\addplot [black]
table {%
4 0.0736173214671656
4 0.0951350552688129
};
\addplot [black]
table {%
3.8875 0.0107763558311707
4.1125 0.0107763558311707
};
\addplot [black]
table {%
3.8875 0.0951350552688129
4.1125 0.0951350552688129
};
\addplot [white]
table {%
0.775 0.108448337124181
1.225 0.108448337124181
};
\addplot [white]
table {%
1.775 0.0540679301026183
2.225 0.0540679301026183
};
\addplot [white]
table {%
2.775 0.0407036069468171
3.225 0.0407036069468171
};
\addplot [white]
table {%
3.775 0.0603474752578764
4.225 0.0603474752578764
};
\end{axis}

\end{tikzpicture}

%% file: sec/05_Discussion.tex
\section{Discussion}
\label{sec:discussion}
Throughout the paper, we have provided a comprehensive comparison of DDS, MQTT, and Zenoh by giving the experiment results and their analysis. The evaluation of latency and throughput within various network setups, including Ethernet, Wi-Fi, and 4G, yielded significant insights into the performance of different middlewares. Besides these insights, the ensuing subsections delve into two other dimensions worth noting—namely, ``CPU Usage'' and ``Security''—to provide a multi-angle understanding of the middleware.

\begin{table}[h]
    \centering
    \caption{Security difference}
    \label{tab:security}
    \begin{tabular}{@{}lccccl@{}}
    \toprule
    \textbf{Security} & \textbf{Mean latency} & \textbf{Throughput} & \textbf{CPU usage} \\ 
    \midrule
    With security & 77.35 ms       & 9701 bytes/sec  &  0.06 \%  \\
    Without security & 72.11 ms    &   9804 bytes/sec &  0.06 \% \\ 
    \bottomrule
    \end{tabular}
\end{table}

\subsection{CPU usage}
The CPU usage of Host 1 as shown in Table~\ref{tab:cpu}, acquired through the ``performance\_test'' tool, provides valuable information about the computational efficiency of each middleware. 

\subsection{Security}
Enabling ROS 2 security affects the performance metrics such as latency and throughput of the ROS\,2 communication channel~\cite{kim2018security}.
The use of Array1k messages to test latency, throughput, and CPU usage underlines the effects of enabling security features. As shown in Table~\ref{tab:security}, the results reveal that activating security measures causes a noticeable increase in mean latency, indicating the additional processing overhead introduced by security mechanisms. Furthermore, there is a decrease in throughput, which aligns with expectations due to the encryption and decryption processes associated with securing message communication. And CPU usage remains relatively unaffected by the security measures.


%% file: sec/06_Conclusion.tex

\section{Conclusion}\label{sec:conclusion}
In this paper, we comprehensively evaluate three networking middlewares used in ROS\,2 system. We compared the latency and throughput of CycloneDDS, Zenoh, and MQTT with ROS Messages under different network setups, including Ethernet, Wi-Fi, and 4G. 
Moreover, we tested the performance of these middlewares on a real robot platform, TurtleBot 4. We set a host and the TurtleBot 4 under the local Wi-Fi network, and use the host to control the TurtleBot 4 to run a square-shape trajectory for 96\,s through different middleware. The trajectories were recorded by the Optitrack MOCAP system. 
Our experimental results reveal that under Ethernet, CycloneDDS has minimal latency and throughput, due to its UDP multicast mechanism. Zenoh exhibits superior performance under Wi-Fi and 4G conditions. On the real robot platform, Zenoh achieves the smallest trajectory drift error.
Our findings that Zenoh performs best are aligned with the survey results from ROS\,2 community that Zenoh was the alternative most suggested by users, and current efforts by the ROS\,2 TSC Meeting to include Zenoh in the next ROS\,2 release.